\documentclass[referee, sn-nature]{sn-jnl}
\usepackage{graphicx}
\usepackage{multirow}
\usepackage{amsmath,amssymb,amsfonts}
\usepackage{amsthm}
\usepackage[title]{appendix}
\usepackage[dvipsnames]{xcolor}
\usepackage{textcomp}
\usepackage{manyfoot}
\usepackage{booktabs}
\usepackage{listings}
\usepackage{caption}
\usepackage{subcaption}
\usepackage{svg}
\usepackage{comment}
\usepackage{mathtools}

\begin{document}
\nolinenumbers
\title[Article Title]{A Concept-based Interpretable Model for the Diagnosis of Choroid Neoplasias using Multimodal Data}

\author[1]{\fnm{Yifan} \sur{Wu}}
\equalcont{These authors contributed equally to this work.}

\author[2]{\fnm{Yang} \sur{Liu}}
\equalcont{These authors contributed equally to this work.}

\author[1]{\fnm{Yue} \sur{Yang}}
\author[1]{\fnm{Michael S.} \sur{Yao}}
\author[3]{\fnm{Wenli}
\sur{Yang}}
\author[3]{\fnm{Xuehui}
\sur{Shi}}
\author[3]{\fnm{Lihong}
\sur{Yang}}
\author[3]{\fnm{Dongjun}
\sur{Li}}
\author[3]{\fnm{Yueming}
\sur{Liu}}
\author[1]{\fnm{James C.} \sur{Gee}}
\author*[3]{\fnm{Xuan} \sur{Yang}}
\email{yangxuan153@126.com}
\author*[3]{\fnm{Wenbin} \sur{Wei}}
\email{weiwenbintr@163.com}
\author*[2]{\fnm{Shi} \sur{Gu}}\email{gus@uestc.edu.cn}

\affil[1]{\orgname{University of Pennsylvania}, \city{Philadelphia}, \state{PA}, \country{US}}
\affil[2]{\orgname{University of Electronic Science and Technology}, \city{Chengdu}, \country{China}}
\affil[3]{\orgname{Beijing Tongren Eye Center, 
Beijing Key Laboratory of Intraocular Tumor Diagnosis and Treatment, 
Beijing Ophthalmology and Visual Sciences Key Lab, 
Beijing Tongren Hospital, Capital Medical University}, \city{Beijing}, \country{China}}

\abstract{
	Diagnosing rare diseases presents a common challenge in clinical practice, necessitating the expertise of specialists for accurate identification. The advent of machine learning offers a promising solution, while the development of such technologies is hindered by the scarcity of data on rare conditions and the demand for models that are both interpretable and trustworthy in a clinical context. Interpretable AI, with its capacity for human-readable outputs, can facilitate validation by clinicians and contribute to medical education. In the current work, we focus on choroid neoplasias—the most prevalent form of eye cancer in adults, albeit rare with 5.1 per million. We built the so-far largest dataset consisting of 750 patients, incorporating three distinct imaging modalities collected from 2004 to 2022. Our work introduces a concept-based interpretable model that distinguishes between three types of choroidal tumors, integrating insights from domain experts via radiological reports. Remarkably, this model not only achieves an F1 score of 0.91, rivaling that of black-box models, but also boosts the diagnostic accuracy of junior doctors by 42\%. This study highlights the significant potential of interpretable machine learning in improving the diagnosis of rare diseases, laying a groundwork for future breakthroughs in medical AI that could tackle a wider array of complex health scenarios.
}

\keywords{Uveal Melanoma, Computer-aided Diagnosis, Interpretable Machine Learning, Multi-modality Classification, Concept Bottleneck Model, Rare Disease Diagnostics}

\maketitle
\newpage

\section*{Introduction}\label{sec1}
Recent advancements in machine learning and deep neural networks have accelerated the development of computer-aided diagnosis (CAD) in the past decade \cite{chan2020computer}. In particular, for common diseases amenable to automated diagnoses with large publicly available datasets, deep learning-based models trained on these data have performed comparably to expert radiologists across a variety of diagnostic tasks. Example applications include analyzing chest X-ray (CXR) screening \cite{johnson2019mimic, lin2023improving}, fundus photography for automated retinopathy screening \cite{gao2023discriminative}, and brain MRIs for tumor and stroke lesion quantification \cite{menze2014multimodal, liew2018large}. However, for most diseases, especially rare diseases, there is usually a lack of high-quality datasets because of the high cost of professional annotations and the incompatibility between clinical and research protocols. In addition, the diagnosis of rare diseases associated with significant patient morbidity and mortality, for example, oncologic diseases, raises high requirements of \textit{interpretable} machine-generated predictions for easy downstream verification by specialized physicians with domain expertise \cite{decherchi2021, molnar2023ai, richens2020improving}. Addressing such challenges is therefore crucial for translating the benefits of CAD technologies for rare disease identification.

In the workup of rare oncologic diseases, physicians use multimodal imaging data collected from different biomarkers and acquisition methods to provide a comprehensive diagnosis for patients. The development of an applicable CAD should also employ such a pipeline. The recent works proposed implementing these pipelines by aligning medical imaging data to the context description \cite{huang2023visual, zhang2023knowledge}, leveraging the representation power of foundation models such as Contrastive Language-Image Pre-training (CLIP) \cite{radford2021learning} and Generative Large Language Model (LLM) \cite{singhal2023large}. While AI methods may help augment the clinical workflows in many common diseases, their applications to rare disease diagnostic tasks are largely unexplored. Notably, when attempting to diagnose rare diseases with artificial intelligence (AI) in particular, there is a need for well-curated data and feasible tools to generate explainable biomarkers aligned with experienced specialists. High-quality data with labeled image-text pairs are required to train sophisticated AI models with sufficient diagnostic performance. Because of the low prevalence of the disease, analytical tools are also needed to offer both interpretable predictions and accurate model predictions to guide clinical management \cite{chae2024strategies, rajpurkar2022ai, varoquaux2022machine}. These practical constraints call for a machine learning paradigm different than existing approaches for diagnosing rare oncologic diseases.

\begin{figure}[ht]
\centering
\includegraphics[width=0.95\textwidth]{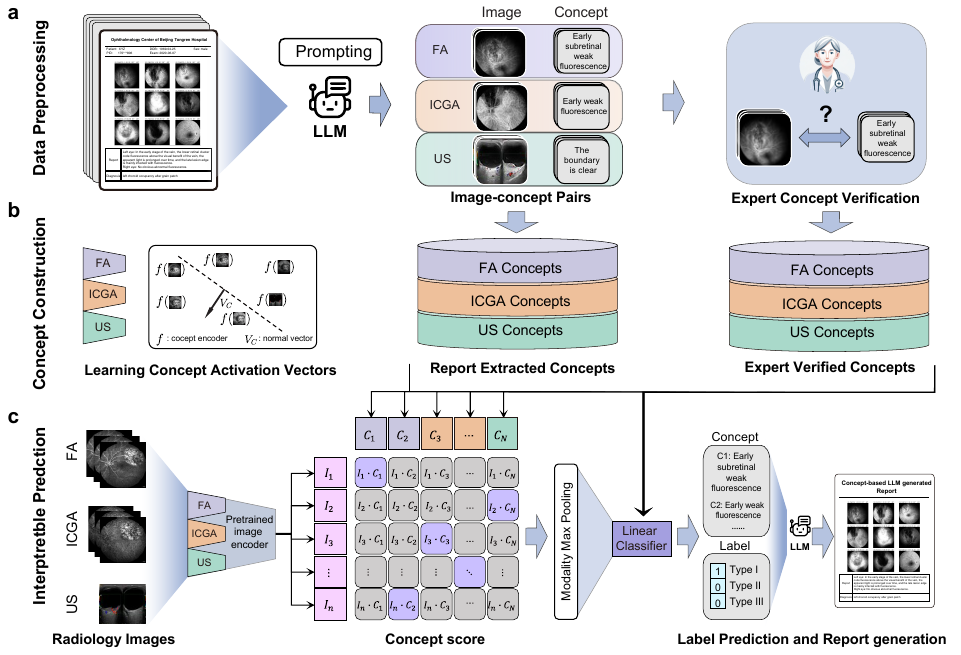}
\caption{\textbf{Overview of the MMCBM workflow.} {(a)} Utilizing a large language model (LLM), concept banks are formulated by extracting image-concept pairs from comprehensive medical reports. Senior experts help examine the faithfulness of the image-concept pairs and make corresponding modifications. {(b)} Based on such pairs, we construct the concept bank by learning concept activation vectors. {(c)} The model's output stage takes a series of images spanning 1 to 3 modalities. A pre-trained image encoder is employed to convert these images into tokenized features. Subsequent calculations produce concept scores. The model then delivers an explainable prediction, spotlighting the diagnostic evidence. Moreover, it crafts an interpretative report, enhancing the transparency of the diagnostic process. }\label{fig1}
\end{figure}

In this work, we discuss our approach on engineering machine learning model architectures specifically designed for rare disease diagnosis. Here, we focus on uveal melanoma, a rare cancer originating from the iris, ciliary body, choroid, or other components of the uveal tract in the eye \cite{jager2020uveal, shields2023metastatic}. While cases of uveal melanoma are rare, with an estimated incidence of $5.1$ per million in the United States, the long-term prognosis is poor due to the high risk of metastasis at the time of diagnosis \cite{singh2011uveal}. Uveal melanomas are frequently missed in routine clinical workups due to their low prevalence in the general population, making few clinicians well-trained in their diagnosis and clinical management \cite{kaliki2015uveal}. Initial diagnosis requires a detailed fundoscopic examination with an expert clinician followed by additional advanced imaging techniques such as ocular ultrasound (US), fluorescein angiography (FA), and indocyanine green angiography (ICGA) for confirmation and prognostication \cite{egan1988epidemiologic, augsburger1990clinical, carvajal2016metastatic, khoja2019meta}. Domain-specific expert physicians specializing in managing uveal melanomas are few and far between, further complicating diagnostic workup \cite{carvajal2016metastatic}. Specifically, in this work, we aim to differentiate between choroidal melanoma, metastatic carcinoma, and hemangioma, all of which occur in the choroid of the fundus and often appear as solitary tumors. These diseases may have similar symptoms in the early stages and overlapping imaging features \cite{mathis2019new}. 
Given the poor prognosis associated with uveal melanomas and the consequent need for timely diagnosis and treatment, it is crucial to have high confidence in a diagnosis of %
choroidal neoplasias prior to definitive intervention. 

To establish such a pipleline for the automated interpretable diagnosis of choroidal neoplasias, we need to address the data, model, and verification questions. First, to enable our work, we collect and release the first well-curated multimodal dataset of uveal oncologic pathologies to train classifier models that accurately differentiate choroidal melanomas from other clinically similar diseases. As far as we know, this dataset is the largest one for choroidal melanomas in the world. 
Next, utilizing this valuable multimodal dataset, we develop a domain knowledge-incorporated model named the Multimodal Medical Concept Bottleneck Model (MMCBM). 
This model predicts interpretable outputs and supports a human-in-the-loop mechanism to accept feedback from domain experts. Furthermore, we find that the extracted pairs of clinical concepts with the corresponding images align well with senior doctors and provide substantial assistance for the junior ones to accurately diagnose choroidal neoplasias. Our methodology leverages the extensive knowledge in clinical reports, offering a pathway toward building interpretable models for diagnosing rare diseases.

\section*{Results}\label{sec2}
\begin{figure}[ht]
\centering
\includegraphics[width=0.95\textwidth]{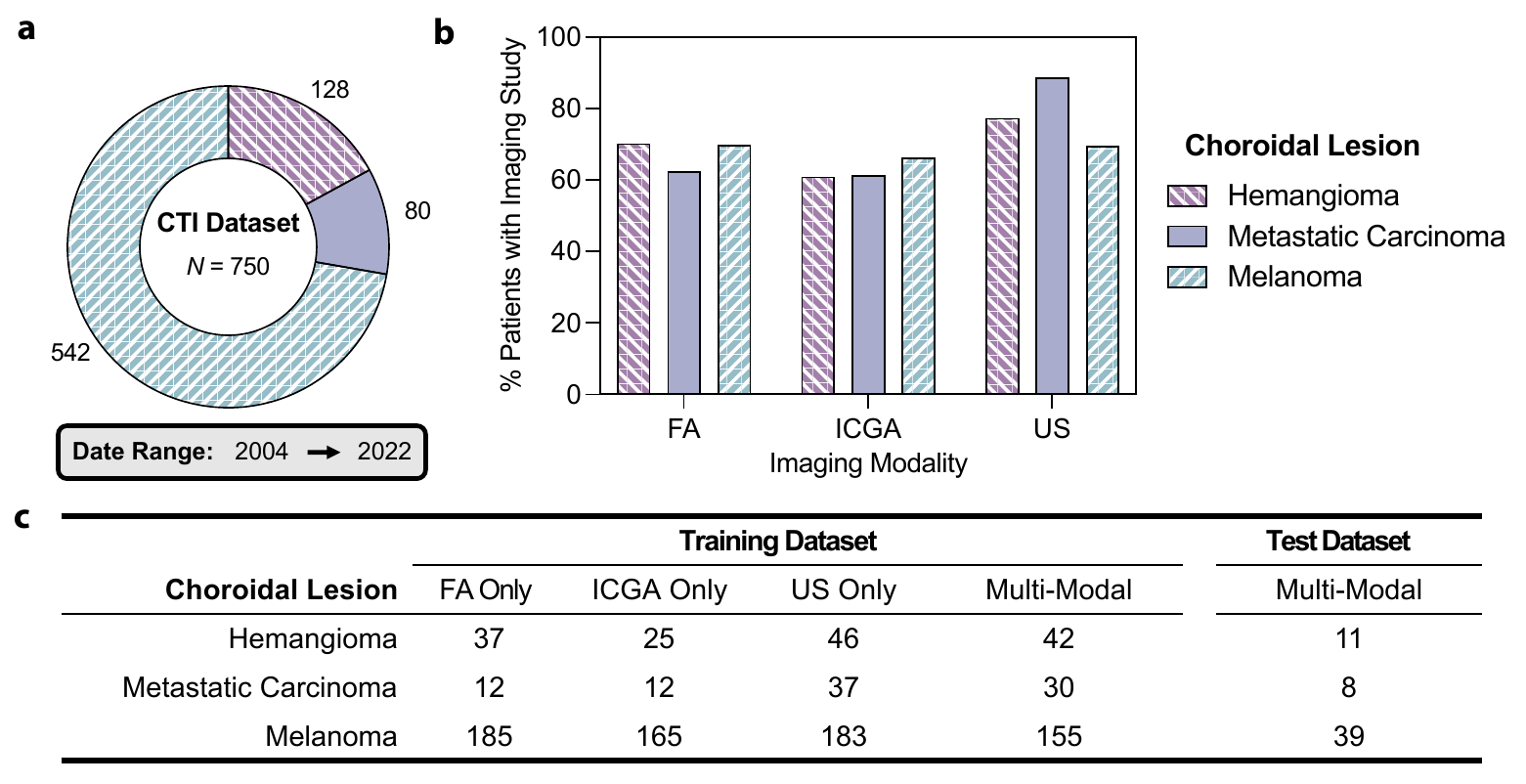}
\caption{\textbf{Statistics of the CTI Dataset. }
{(a)} The CTI dataset is composed of 750 patients: 542 with melanoma, 128 with hemangioma, and 80 with metastatic carcinoma, collected from 2004 to 2022.
{(b)} Proportions of patients with hemangioma, metastatic carcinoma, and melanoma imaged by Fluorescein Angiography (FA), Indocyanine Green Angiography (ICGA), and Ultrasound (US). 
{(c)} Split of imaging studies in the training and test datasets across various imaging modalities: 20\% of the Multi-Modal data (MM), representing patients imaged with all three modalities, is set aside for testing. The remaining 80\% of MM and all non-MM data are allocated for training using 5-fold cross-validation. }
\label{fig2}
\end{figure}

\label{sec2.1}
\noindent\textbf{Dataset Description}\\ 
To support the development of interpretable models for diagnosing choroidal tumors, we built the Choroid Tri-Modal Imaging (CTI) dataset, a multimodal annotated collection of medical images from Beijing Tongren Hospital (2004-2022), encompassing Fluorescence Angiography (FA), Indocyanine Green Angiography (ICGA), and Ocular Ultrasound (US). This dataset, approved by the Ethics Committee of Beijing Tongren Hospital, includes images from patients diagnosed with benign hemangioma, secondary metastatic carcinoma to the eye, or primary choroidal melanoma. The CTI dataset (Fig.~\ref{fig2}) consists of 542 patients with choroidal melanoma (FA: 379, ICGA: 359, US: 377), 128 patients with choroidal hemangioma (FA: 90, ICGA: 78, US: 99), and 80 patients with choroidal metastatic carcinoma (FA: 50, ICGA: 49, US: 71). The numbers indicate the quantity of imaging studies for each specific modality. Note that not every patient has images across all modalities. We refer to the subset where patients have all three modalities as Multi-Modal (MM) data and reserve 20\% of this MM data for testing purposes throughout this study. In the MM data training split, 97 patients have anonymized reports for all three modalities, describing the radiological features observed in the images. \\ 

\noindent\textbf{Baseline Blackbox Model}\\
We first sought to build black-box machine learning models. 
Taking inspiration from recent success in natural image processing with efficient network design \cite{tan2019efficientnet}, our baseline black-box model is composed of three separate modality-specific encoders trained to encode corresponding imaging study inputs into intermediate lower-dimensional representations. The encoder output (or \textit{outputs}, if multiple imaging studies of different modalities are available for a given patient) is then passed to an attention pooling block \cite{safari2020self} 
and final dense layer to yield the final classification prediction. We refer to this model architecture as the \emph{Pre-Trained Multimodal Classifier}. Our baseline model performs accurately across different input image modalities, validating the feasibility of deep-learning models for this clinical problem. Using FA imaging studies alone, the pre-trained classifier achieved an $F_1$ score of 78.3\% (95\% CI: 74.0 - 81.7\%); using ICGA studies alone, it achieved an $F_1$ score of 85.9\% (95\% CI: 83.7 - 88.2\%); and using US studies alone, it achieved an $F_1$ score of 72.1\% (95\% CI: 67.1 - 76.7\%). When using all three imaging studies together, the baseline classifier attained an $F_1$ score of 89.2\% (95\% CI: 87.9 - 90.6\%). The metrics of accuracy, precision, and recall are comprehensively presented in Table \ref{tab:taba0}.
These results support using a multimodal approach to train models that are more accurate than any individual imaging studies as input alone. However, while the Pre-Trained Multimodal Classifier demonstrates impressive performance, it is impossible to reliably interrogate the model's predictions that are easy for human experts to interpret \textemdash a key limitation or existing black-box approaches. \\

\noindent\textbf{Trustworthy Interpretable Framework}\\ 
The lack of interpretability in the baseline pre-trained classification model is a common trait of many modern AI tools. To address the need for trustworthiness in medical diagnostics, we sought to engineer a framework with interpretability baked into the model design. Our approach, referred to as the {Multimodal Medical Concept Bottleneck Model} ({MMCBM}), is a task-agnostic framework designed for high-stakes applications where human-in-the-loop or subsequent human verification is critical. Our key insight is that we leverage prior knowledge from domain experts to align the intermediate representations of input images by the model as representations that human proxies can easily understand. In this way, model predictions can be easily interpreted as activations and linear combinations of these representations. These representations can be visual patterns or findings that clinical experts consider the evidence for making diagnoses and are used as educational guidelines in a natural language format. We refer to these representations as concepts. \\

\noindent\textbf{Concept Construction and Grounding}\\ 
Using medical reports as the knowledge database, we prompt GPT-4 \cite{openai2023gpt4} for concept extraction and construct a concept bank filled with phrases related to imaging findings of choroidal tumors. For instance, a description in a fluorescein angiography (FA) report states, ``In the venous phase, a clustered hypofluorescence under the subretinal can be seen in the temporal part of the macula. Fluorescence increases with time, and lesions are dominated by fluorescent staining at the late stage". The extracted concepts for this FA study are ``Clustered Hypofluorescence During Venous Phase", ``Globally Increasing Fluorescence Intensity", and ``Late-Stage Staining". After extracting concepts from the reports of 97 patients, we use GPT-4 to aggregate semantically similar concepts, ensuring each concept's uniqueness and relevance. The final concept bank consists of 47 concepts for FA, 30 for ICGA, and 26 for US, with an average of 3 concepts for FA, 2 for ICGA, and 5 for US per patient. The comprehensive list of all $N=103$ concepts is presented in Table \ref{tab:taba1}. To validate that the concepts extracted by the LLM accurately represented real-world clinical reasoning, two senior ophthalmologists specializing in diagnosing and managing choroidal tumors at Beijing Tongren Hospital were asked to verify and amend the concepts. Quantitatively, the initial concept bank constructed by GPT-4 was assessed to be reasonable and relevant, necessitating only minor modifications ({Fig.~{\ref{figa8}}}): 5 concepts were removed, and 8 new ones added to the FA category, 4 additional concepts were added to ICGA, and there were no changes to the US category.

\begin{figure}
    \centering
    \includegraphics[width=0.95\textwidth]{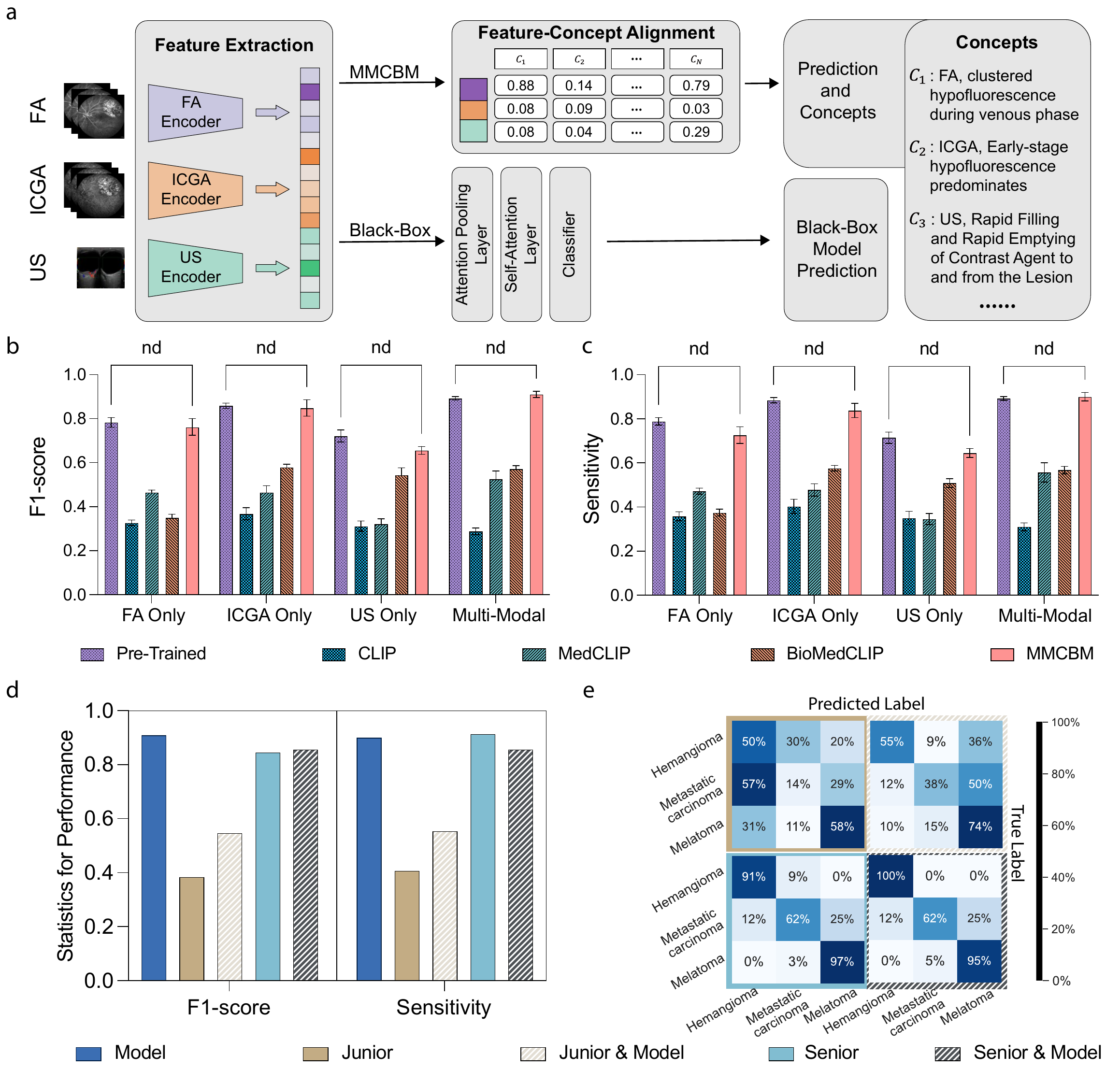}
    \caption{\textbf{Multimodal Medical Concept Bottleneck Model (MMCBM).} Black-box models such as the pre-trained classifier learn directly from the encoded image features and output a single model prediction without any insight as to how the prediction was computed. In contrast, the MMCBM shown in {(a)} instead represents encoded features by their alignment with key medical concepts derived from domain experts. This allows MMCBM to return not only its prediction but also the top-$k$ activated concepts that best describe the input images, giving insight into how the model arrived at its prediction. Comparing both the classification {(b)} accuracy and {(c)} sensitivity of the models, there is no statistically significant difference between black-box models and MMCBM across all sets of imaging inputs. MMCBM concepts also outperform features derived from CLIP-based models, highlighting the importance of source prior knowledge from domain experts. (d) Performance Benchmark with Human Evaluators: A comparison of our model's performance against junior and senior doctors. After presenting them with the model's predicted concepts, they conducted a subsequent assessment, enabling us to document and compare performance metrics. (e) Confusion Matrix for Human Evaluators w/wo Concepts. The matrices correspond to Junior, Junior \& Model, Senior, and Senior \& Model groups and are with corresponding colors.}
    \label{fig3}
\end{figure}

To ground concepts to feature embeddings, we employed SVMs for binary classification per concept. We used image representations from a pre-trained model as input and labels derived from the concept construction process—images with assigned concepts as positive samples, and all others as negative. The classification hyperplane vector from each SVM serves as the concept's representation, which we refer to as concept activation vectors (CAVs) \cite{kim2018interpretability}. Subsequently, in MMCBM, an image is projected into the space of concepts to estimate how much an input image aligns with a modality-specific concept. The alignment scores are then input into a linear classifier to predict the relative probabilities of each of the three targeted choroidal diseases. The Fig.~\ref{fig3}a details this process further, highlighting the top-$k$ concepts derived from concept scores to explain the model's predictions. \\

\noindent\textbf{Noninferior Accuracy of MMCBM to Black-Box Model}\\ 
A common critique of interpretable machine learning models is that enforcing priors on the model, such as requiring input images to align with concept activation vectors, is equivalent to adding additional regularization to the hypothesis space \cite{koh2020concept}. Such regularization may adversely impact the performance of trained models \cite{yuksekgonul2022post, yang2023language}. To this end, we sought to evaluate the classification performance of our MMCBM model against the black-box pre-trained multimodal classifier baseline (Fig.~\ref{fig3}). On the MM testing dataset, MMCBM achieved an overall classification $F_1$ score of 91.0\% (95\% CI: 88.2 - 93.4\%), which is comparable with the performance of the baseline black-box model (89.2\%; 95\% CI: 87.9 - 90.6\%). Contrary to concerns, the MMCBM framework outperformed the pre-trained model. We attribute this improvement to the fact that by adding interpretable regularization, the framework mitigates the issue of class imbalance in the data. Additionally, comparing classifier performance across unimodal imaging inputs revealed no statistically significant differences in classification metrics (Table~\ref{tab:taba0}). This indicates that our MMCBM framework matches the performance of black-box approaches in automating the diagnosis of rare choroidal tumors, according to clinically relevant measures. \\

\noindent\textbf{Integration of MMCBM in Clinical Workflows}\\
We have shown that the MMCBM is a framework that leverages prior knowledge from domain experts to represent input data aligned with interpretable concepts. However, it remains unknown whether our framework can provide real-world utility in augmenting existing clinical workflows. To investigate the applications of MMCBM in clinical practice, we recruited 8 doctors from Beijing Tongren Hospital, 2 senior ophthalmologists specializing in the diagnosis and management of choroidal melanomas, and 6 resident ophthalmologists in training. We assessed the diagnostic performance of ophthalmologists alone against ophthalmologists with our trained MMCBM model. 
The ophthalmologists leveraging our MMCBM model for diagnostic workflow augmentation have access to the top $k$ activated concepts from the MMCBM concept bank and can adjust the confidence scores of the concepts based on their judgment. 
This human-in-the-loop interactive feature improves the practical utility of MMCBM in clinical decision-making, fostering a more collaborative and accurate diagnostic process. For the 6 junior ophthalmologists, the average accuracy is 51.9\%, precision 40.5\%, recall 40.9\%, and $F_1$ score 38.5\%; with the aid of our MMCBM model, their accuracy improves to 65.5\%, precision to 54.3\%, recall to 55.5\%, and $F_1$ score to 54.7\% ({Fig.~\ref{fig3}d, e}). The 2 senior ophthalmologists demonstrate a high diagnostic accuracy at baseline of 91.4\%, precision of 85.8\%, recall of 83.6\%, and $F_1$ score of 84.6\%. When augmented with the model’s predictions, their performance remains relatively unchanged with an accuracy of 91.4\%, precision of 86\%, recall of 85.8\%, and $F_1$ score of 85.7\%. In particular, the use of MMCBM improves junior doctors' performance by 42\% on the $F_1$ score. These results not only validate the quality and precision of the predicted concepts of our MMCBM model but also highlight our model's ability to serve educational purposes by improving the diagnostic accuracy of less experienced doctors for complex and rare diseases. \\

\noindent\textbf{Comparison between MMCBM and Alternative Feature Embedding Methods}\\
Given the recent progression of cross-modality foundation models, it may be possible to leverage existing feature embedding models trained on extensive corpora of medical information to represent input ocular imaging data and concepts. This approach might offer greater generalizability and require less effort than our MMCBM setup. To evaluate this alternative framework, we compared our concept embedding procedure and image feature extraction with those using Contrastive Language-Image Pre-training (CLIP) \cite{radford2021learning} and its biomedical variants, including MedCLIP \cite{wang2022medclip} and BioMedCLIP \cite{zhang2023large}, which are specifically fine-tuned for medical data. Briefly, MedCLIP was fine-tuned on multiple Chest X-ray datasets, while BioMedCLIP underwent fine-tuning on 15 million figure-caption pairs extracted from biomedical research articles in PubMed Central. Overall, we can see that all CLIP-based frameworks perform inferiorly with statistical significance compared to our CAV-based feature extraction method employed in our MMCBM framework, which achieves a classification $F_1$ of 91.0\% (95\% CI: 88.2 - 93.4\%) on multimodal image inputs. In addition, methods fine-tuned on specialized medical datasets\textemdash such as MedCLIP and BioMedCLIP\textemdash outperform the generic CLIP model as feature extractors for choroidal disease diagnosis for both multimodal and unimodal image inputs ({Fig.~\ref{fig3}b,c}, MedCLIP: 52.5\% (95\% CI: 47.2 - 59.6\%), BioMedCLIP: 57.2\% (95\% CI: 54.7 - 59.6\%), CLIP: 28.8\% (95\% CI: 26.2 - 31.3\%)). The analysis of unimodal input results and additional classification metrics further aligns with these findings. Specifically, embedding model inputs with expertise-curated knowledge significantly outperforms the use of general domain knowledge. These observations highlight the necessity for fine-tuning and domain-specific adaptation or embedding images and texts in medical applications. Furthermore, they affirm the efficacy of our MMCBM as a viable and effective means to achieve model interpretability without compromising algorithmic performance. \\
 
\begin{figure}[htbp]
\centering
    \includegraphics[width=0.95\textwidth]{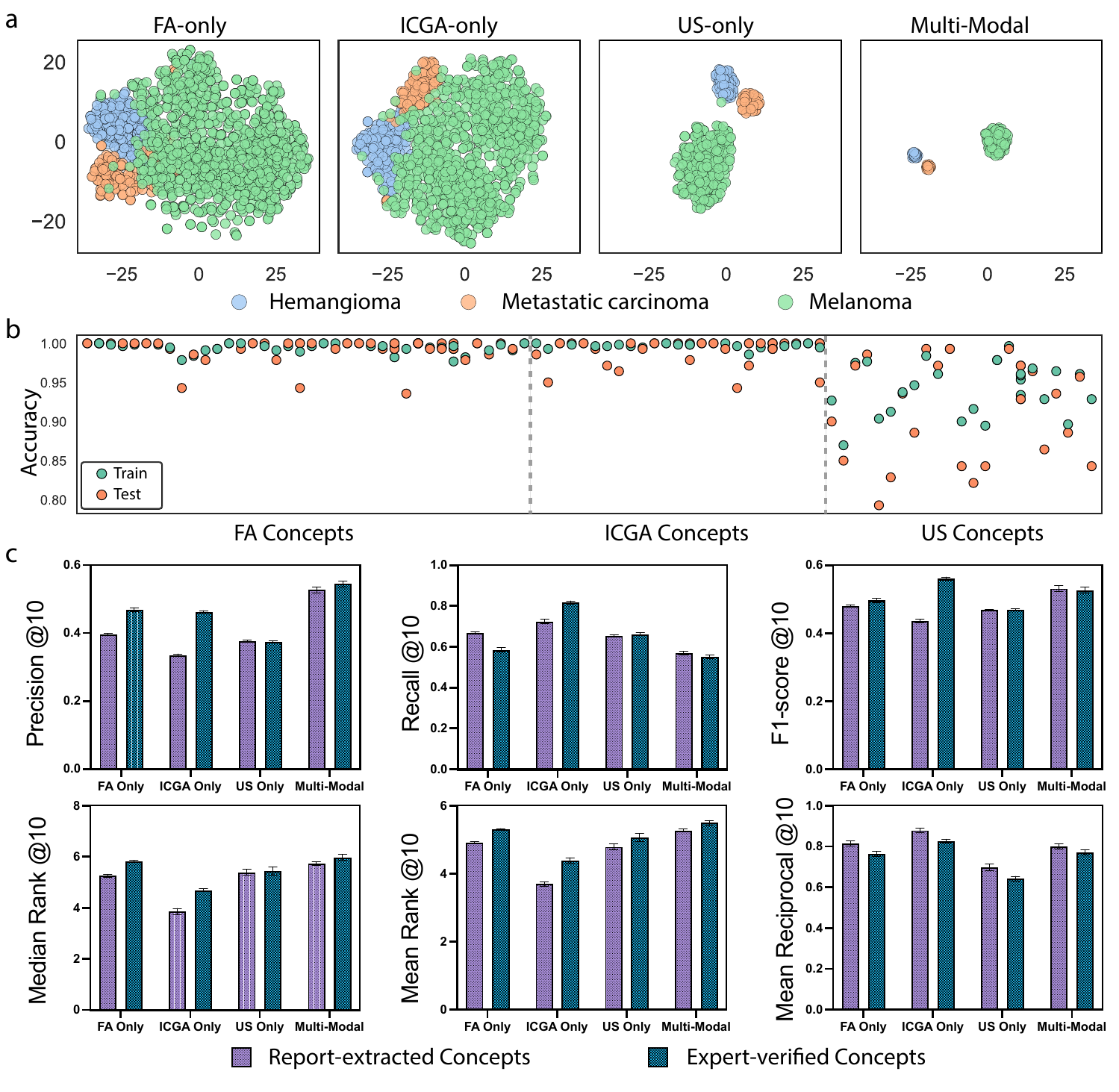}
    \caption{\texttt{|} Comparative Human Evaluation and Model Insights. (a) Embedding Visualizations via t-SNE: This offers a graphical representation of embeddings from the trio of pretrained encoders. Notably, the fused MM embeddings are processed through the attention-pooling mechanism. (b) Accuracy of SVMs in generating concept banks using Concept Activation Vectors (CAVs). (c) Metrics of predicted Top-k concepts on test dataset with k = 10. This evaluation includes precision@k, recall@k, and F1@k, as well as mean rank@k, median rank@k, and mean reciprocal rank@k. 
}
    \label{fig4}
\end{figure}
\noindent\textbf{Evaluation of Image-Concept Alignment}\\
Our MMCBM demonstrates classification performance on par with state-of-the-art black-box models and offers interpretable insights into final model outputs. While the classification performance may benefit from the black-box models, the quality of model interpretability highly depends on the quality of the prior knowledge used to construct the MMCBM concept bank, the quality of the image and concept embedding functions, and the quality of image-concept alignment. Thus, this section evaluates our model's interpretability quality from these three aspects.

First, we evaluated the MMCBM feature representations and their accuracy in describing input images. Model representations for each of FA, ICGA, and US imaging studies were computed by the respective MMCBM encoders before leveraging t-SNE \cite{van2008visualizing} dimensionality reduction techniques to visualize the complex feature landscapes from our multimodal dataset ({Fig.~\ref{fig4}a}). We observe distinct clusters corresponding to, hemangioma, metastatic carcinoma, and melanoma, indicating effective class separation by the MMCBM encoders. Qualitatively, the clusters corresponding to multimodal data inputs appear more cohesive and less dispersed, suggesting that integrating multi-modal inputs may improve the separability of the different class representations in this representation space. This enhanced clustering density may contribute to the improved discriminative performance of our multimodal MMCBM models in contrast to models with only unimodal inputs accessible. 

Next, we evaluated the quality of the MMCBM concept representation and image-concept alignment by examining the accuracy of Support Vector Machine (SVM) classifiers employed in generating concept vectors for each medical concept. A high SVM accuracy score indicates a concept's representational effectiveness and consistent presence across the dataset. According to this metric, FA and ICGA concepts achieve high accuracy across the board ({Fig.~\ref{fig4}b}), with accuracy on test data exceeding 90\% for all concepts. This suggests that concepts derived from FA and ICGA are well-represented and aligned with the input images. In contrast, though less accurate, the accuracy scores for US-based concepts are still higher than 80\% for all concepts, indicating satisfactory performance. This indicates that distinguishing diseases from ultrasound images alone may be more challenging. Specific details of the individual concepts and their corresponding accuracies are detailed in Table \ref{tab:taba1}. 

To further assess the quality of MMCBM concept-based interpretability, we examined how well the model concepts align with ophthalmologist annotations. We selected the top-$k$ concepts predicted by MMCBM for each patient in the multimodal testing dataset. The {Fig.~\ref{fig4}c} offers a comprehensive quantification of the model's alignment with expert annotations according to key performance metrics: Precision@$k$, Recall@$k$, F1@$k$, Median-Rank@$k$, Mean-Rank@$k$, and Mean-Reciprocal-Rank-(MRR)@$k$, with $k$ = 10. We compared two setups of concept banks: the report-extracted and the expert-verified. We found that report-extracted concepts achieved Precision@10 = 0.53 and Recall@10 = 0.57, similar to expert-verified concepts (Precision@10 = 0.54, Recall@10 = 0.55). It is worth noting that expert-verified concepts yielded better alignment with expert annotations, suggesting that human intervention in the verification process improves the concept bank's ability to capture domain knowledge. Our analysis demonstrates that the MMCBM model concepts extracted from reports closely match the performance of expert-verified annotations across various metrics. This suggests that report-extracted concepts achieve interpretability comparable to expert-verified concepts, negating the need for time-intensive expert annotation while effectively capturing the salient clinical features of interest to ophthalmologists. \\

\begin{figure}[htbp]
\centering
\includegraphics[width=0.90\textwidth]{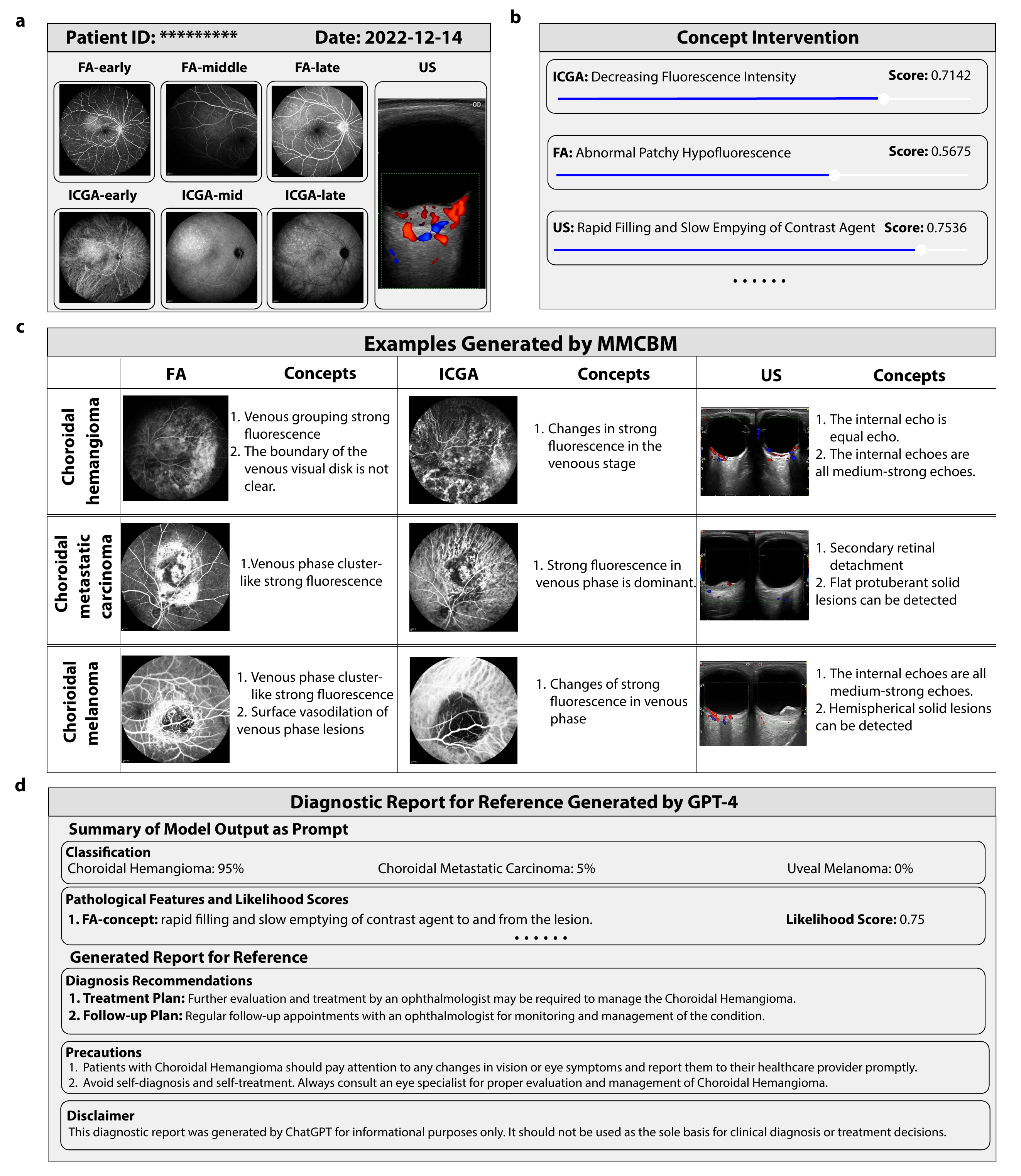}
\caption{\textbf{Demo of Human Interactive Interface.} We make a website to facilitate the user interactive study with ophthalmologists. (a) Image Display Panel: as FA and ICGA imaging span various time frames, ophthalmologists pinpoint images from early, medium, and late phases for accurate classification. (b) Interventions interface on concept bottleneck: a panel that allows adjustment of the concept scores to refine the final prediction. (c) Visual Emphasis on Bottlenecks: a curated selection of representative cases processed by the model, highlighting the top-k concepts prioritized by their attention scores in the weight matrix displayed across three distinct tumor classes. (d) Diagnostic Reporting in Action: an example of a diagnostic report formulated by ChatGPT during the testing phase. The input to ChatGPT includes the predicted top-k concepts combined with patient-specific details, highlighting the model's capability to produce interpretable diagnoses. 
}\label{fig5}
\end{figure}

\noindent\textbf{Demo of Human-Model Interaction}\\
To exemplify a practical engineered system for enabling human-model interactions, we make available our website (\href{https://xxxxxxxxxxxxxxxxx}{https://xxxxxxxxxxxxxxxxx}) used for this user-based study with the eight ophthalmologists. Our website provides a user-friendly online interface for concept bank verification and predication evaluation. The annotation system allows ophthalmologists to upload images and annotate them with clinically meaningful concepts ({Fig.~\ref{fig5}a}) or verify images along with report extracted concepts. The prediction system can accept FA, ICGA, and/or US images, and use them to output imaging concepts with confidence scores ({Fig. \ref{fig5}b}). In instances where MMCBM may produce erroneous concept predictions, clinicians can adjust the confidence scores of individual concepts. Such adjustments can refine and correct the model predictions, aligning them more closely with clinical findings. This feature of human intervention significantly improves the practical utility of MMCBMs in clinical decision-making, fostering a more collaborative and accurate diagnostic process. The {Fig.~\ref{fig5}c} displays several examples generated by MMCBM, including a curated selection of representative cases processed by the model. Finally, given the model outputs, a basic diagnostic report can be generated by leveraging LLMs to interpret the MMCBM outputs and concept activations ({Fig. \ref{fig5}d}). It highlights the top-$k$ activated concepts and presents the final generated diagnostic report.
The report generation prompt example can also be found in {Fig.~\ref{figa6-rg}}. \\

In summary, our results highlight the MMCBM model as a promising tool for clinical decision support. While the model's predictions are accurate on their own, they are most effective when combined with human expertise, offering the most comprehensive diagnostic performance and underscoring the potential of AI-assisted diagnostics.

\section*{Discussion}\label{sec3}

In this work, we establish the Multimodal Medical Concept Bottleneck Model (MMCBM) as a novel approach for the interpretable diagnosis of rare choroid tumors. To facilitate the application of advanced machine learning techniques, we initially tackled the significant challenge of scarce comprehensive training data by curating the Choroidal Tri-Modal Imaging clinical dataset. This dataset, which includes image data of Fluorescein Angiography (FA), Indocyanine Green Angiography (ICGA), and Ultrasound (US) with associated radiology reports, to our knowledge, is the largest dataset of choroidal melanoma. Based on this dataset, our MMCBM maintains the accuracy of prior ``black-box" models and introduces interpretability through the concept bottleneck model. Furthermore, by incorporating the explainable MMCBM into the diagnostic workflow, our model significantly enhances the performance of junior ophthalmologists.

Unlike traditional methods for explainable AI, which often rely on saliency maps \cite{selvaraju2017grad, wu2020interpretable, tjoa2020survey} to highlight important spatial attributes, our approach aligns more closely with clinical practice, which is inherently finding-based. Clinicians identify a range of descriptive visual features, including textual elements, contrast, shape, and dynamic changes, that extend beyond mere location. Traditional approaches to incorporating this additional information and prior knowledge typically require either expensive labeling or sophisticated network infrastructure designs to integrate clinical insights, thereby limiting the generalized utility of explainable AI tools. Our introduction of ``concepts" addresses this gap by providing human-comprehensible descriptions that facilitate intervention in the diagnostic process. This yields a twofold benefit: it simplifies the alignment between domain knowledge in clinical practice and the representational power of neural networks, and it proves immensely beneficial for junior doctors, who may lack experience in finding identification and risk over-reliance on AI outputs \cite{kostick2022ai}.

Moreover, recent advancements in vision and natural language processing, such as Large Language Models (LLMs) and Contrastive Language–Image Pre-training (CLIP), have paved new pathways for research into interpretable diagnostic systems. However, for rare diseases like choroidal melanoma, the scarcity of paired image-text knowledge on the internet presents a significant challenge to the reliability of these models' reasoning capabilities, as evidenced in Fig.~\ref{fig3}. While professional annotation of high-quality data can mitigate this issue, further data access and expertise challenges remain \cite{wang2022medclip}, especially for rare diseases. Our concept-based multimodal model circumvents these challenges by utilizing LLMs to process texts without necessitating detailed labeling of image features. The model's predictive and interpretive power stems from integrating the pre-trained model with the extracted relationship between reports and images. This approach mitigates the data scarcity issue for rare diseases in recent foundation models, avoiding the need for extensive labeling efforts in medical AI preparation, thus making the design extendable to other rare diseases.

In the realism of AI-aid medical diagnosis, particularly for the detection and intervention of serious diseases like the choroid neoplasias we considered in the current work, ethical considerations are of critical importance \cite{char2020identifying}. Our methodology, which enables the human-in-the-loop (HITL) mechanism, helps address this issue by aligning human expertise with AI diagnosis. Specifically, by actively involving domain experts in the training and validation phases of AI model development, we not only ensure that the AI's diagnostic concepts are vetted by experienced clinicians but also provide feasible constraints of the degree of AI intervention, thereby reducing the risk of hallucination that could arise from sole reliance on AI. This approach may foster trust among clinicians and patients in AI-assisted medical decisions. The inclusion of HITL integration in our AI models aligns with ethical guidelines for AI in healthcare, emphasizing the safeguarding of patient dignity and privacy. As we advance the frontiers of medical AI, it is crucial to maintain a balanced synergy between technological innovation and ethical responsibility, ensuring that AI serves as a supportive tool rather than a replacement for the nuanced judgment of medical professionals.

While our results presented in the study are promising, it nonetheless has limitations. Firstly, the multi-modal data can be noisy due to inconsistencies in image acquisition and labeling. Our filtering pipeline improves data quality, as confirmed by human expert evaluation, but it still requires careful oversight to ensure that the MMCBM concepts are well aligned with prior knowledge. Secondly, as with any application of machine learning in healthcare, the clinical implementation of such models requires rigorous validation through prospective studies and randomized clinical trials. Collaboration with regulatory bodies will ensure these diagnostic tools meet safety, efficacy, and equity standards. We hope to explore the capacity of MMCBMs and other interpretable models to meet these standards in future work.

In summary, the development of MMCBMs marks a significant advancement toward achieving interpretable and reliable diagnoses within the healthcare domain. As efforts to refine and incorporate these models into clinical workflows progress, it is imperative to carefully consider the ethical and regulatory dimensions to ensure that these innovations enhance patient outcomes without compromising the standards of care or jeopardizing patient safety. This work delineates a promising avenue for applying artificial intelligence in the nuanced and critical field of diagnosing rare diseases, offering a blueprint for future explorations in this vital area of medical research.

\section*{Materials and Methods}\label{sec4}
\noindent\textbf{Dataset Collection and Ethics Statement.} The patient data in the CTI dataset were collected at Beijing Tongren Hospital from January 2004 to December 2022 (Approval No. TRECKY2018-056-GZ(2022)-07). To our knowledge, it is the largest clinical database containing multimodal data from patients with choroidal melanoma and other closely related ocular pathologies. This extensive database contains diagnostic and pathological data of patients with choroidal diseases. The database includes a total of 925 cases, which comprise 161 cases of choroidal hemangioma, 82 cases of choroidal metastatic carcinoma, and 682 cases of choroidal melanoma. The image collection includes three types of radiological images: fluorescein angiography (FA), indocyanine green angiography (ICGA), and Doppler ultrasound images (US). Each patient has one or more modalities of images. The FA and ICGA images, being time-series, were captured from three angles: 30, 55, and 102 degrees. The US images include two types: B-mode ultrasound and color Doppler ultrasound. Medical professionals have thoroughly reviewed the data-cleaning process to ensure its integrity and clinical relevance. For FA and ICGA modalities, we ignored the shooting angle and categorized the FA and ICGA images into three periods\textemdash early, middle, and late\textemdash in alignment with existing clinical diagnostic recommendations. The time frames for these periods are as follows: ICGA (Early: less than 5 minutes; Middle: between 5 and 20 minutes; and Late: at least 20 minutes) and FA (Early: less than 5 minutes, Middle: between 5 and 10 minutes, Late: at least 10 minutes). We selected binocular color Doppler images containing blood flow information for the US modality. Finally, the cleaned dataset includes a total of 750 cases, which comprises 128 cases of choroidal hemangioma, 80 cases of choroidal metastatic carcinoma, and 542 cases of choroidal melanoma. There are 53 patients with choroidal hemangioma, 38 patients with choroidal metastatic carcinoma, and 194 patients with choroidal melanoma with all three imaging modalities, which we refer to as multi-modal (MM) data. Additionally, 97 cases have clinical diagnostic reports that describe the radiological features observed in the FA, ICGA, and US images. Informed consent was obtained from all patients whose anonymized and de-identified data is included in the dataset. Per the Declaration of Helsinki 2000, the collecting organization obtained written informed consent from the patients. \\

\noindent\textbf{Data Splitting.} To optimize data utilization and establish reliable evaluation indicators, we initially allocated 20\% of patients with all three imaging studies as the test set and performed 5-fold cross-validation at the patient level on the remaining data. Specifically, the remaining data is split into five folds based on each pathology and modality. Data augmentation techniques, including random horizontal flipping, random rotating, and random zooming, were applied during training. To build the multi-modal concept banks, we used 97 diagnosis reports, comprising 39 cases of choroidal hemangioma, 18 of choroidal metastatic carcinoma, and 40 of choroidal melanoma. Each report included three modal images and prompted GPT-4 to extract relevant medical concepts from reports. The prompts are detailed in {Fig.~\ref{figa6}}, and the extracted concepts are in {Table. \ref{tab:taba1}}. \\

\noindent\textbf{Model Training.} Consider a training dataset $\mathcal{D}_{train} = \left\{(\mathbf{x}, \mathbf{r}, y)\right\}$ comprising image-report pairs, where $\mathbf{x}\in\mathcal{X}$ represents a fundus image (of any imaging modality), $\mathbf{r}\in\mathcal{R}$ is the clinical patient report collected by doctors, $y\in\mathcal{Y}\coloneqq\{\text{hemangioma}, \text{carcinoma}, \text{melanoma}\}$ is the corresponding disease label. We utilize GPT-4 to analyze the reports and extract relevant concepts, represented as a function $\mathrm{LLM}: \mathcal{R}\rightarrow \mathcal{C}$ where $\mathcal{C}$ is the space of concepts. 
We can then prompt GPT-4 to combine concepts with the same semantic meaning, resulting in a compressed representation of $N$ concepts $\mathcal{C}=\{c_{1}, c_{2}, ..., c_{N}\}$.
Using a pre-trained multi-modality backbone $\phi: \mathcal{X}\rightarrow \mathcal{Z}$ capable of mapping different modality images into a shared feature space, we can generate bottleneck embeddings to establish a concept bank, denoted as $\mathcal{Z}_{\mathcal{C}} \in \mathbb{R}^{N \times d}$, where $N$ is the number of concepts and $d$ the size of the embedding space of $\phi$. Row $i$ of the two-dimensional matrix $\mathcal{Z}_\mathcal{C}$ represents the learned representation of the $i$th concept $c_i$ obtained through Concept Activation Vectors (CAVs) \cite{kim2018interpretability}. MMCBM generates a prediction $\hat{y} = g\left(\text{sim}\left(\phi(x), \mathcal{Z}_{\mathcal{C}}\right)\right)$. The function $\text{sim}: \mathbb{R}^{d} \rightarrow \mathbb{R}^{N}$ computes the concept scores by calculating the similarities between image features and each element of the concept bank $\mathcal{Z}_\mathcal{C}$. The function $g: \mathbb{R}^{N} \rightarrow \mathcal{Y}$ predicts the final label based on the concept scores, serving as an interpretable predictor. To learn the MMCBM, we solve the following problem:
\begin{equation}
    \min_{g} \mathop{\mathbb{E}}_{(\mathbf{x}, c, y) \sim \mathcal{D}} \mathcal{L}\left[g\left(\text{sim}\left(\phi(x), \mathcal{Z}_\mathcal{C}\right)\right), y\right]
\end{equation}
where $\phi(x)$ is the projection onto the concept space and $\mathcal{L}$ the cross-entropy loss. To ensure that final predictions $\hat{y}$ can be easily derived from inputs $\text{sim}(\phi(x), \mathcal{Z}_\mathcal{C})$, we model $g$ as a linear classifier.\\

\noindent\textbf{Evaluation of Model Performance.} Using a 5-fold cross-validation framework, we report the macro-averaged metrics accuracy, precision, recall, and F1 score, which considers both precision and recall while addressing potential class imbalances. In addition to these traditional classification metrics, we also focused on interpretability metrics such as Precision@$k$, Recall@$k$, Mean Rank@$k$, and Median Rank@$k$. Precision@$k$  measures how many of the top-$k$ identified concepts were right compared with the annotated ground truth.  Recall@$k$ evaluates the ratio of correct concepts in the first k predictions to all correct concepts for the patient. F1@$k$ is the harmonic mean of Precision@$k$ and Recall@$k$. Mean Rank@$k$ and Median Rank@$k$ indicate the average ranking position of the correct concept; lower scores are better. 

\section*{Acknowledgements}
S.G. is supported by NSFC Key Program 62236009, Shenzhen Fundamental Research Program (General Program) JCYJ 20210324140807019, NSFC General Program 61876032, and Key Laboratory of Data Intelligence and Cognitive Computing, Longhua District, Shenzhen. X.Y. and WB.W. are supported by NSFC 82220108017, 82141128, 82002883, the Capital Health Research and Development of Special 2024-1-2052, Science \& Technology Project of Beijing Municipal Science \& Technology Commission Z201100005520045, Sanming Project of Medicine in Shenzhen No. SZSM202311018.

\backmatter

\bibliography{sn-bibliography}

\newpage
\begin{appendices}
\section{}
\subsection{Supplementary Results}\label{secA2}
\begin{figure}[htbp]
\centering
\includegraphics[width=\textwidth]{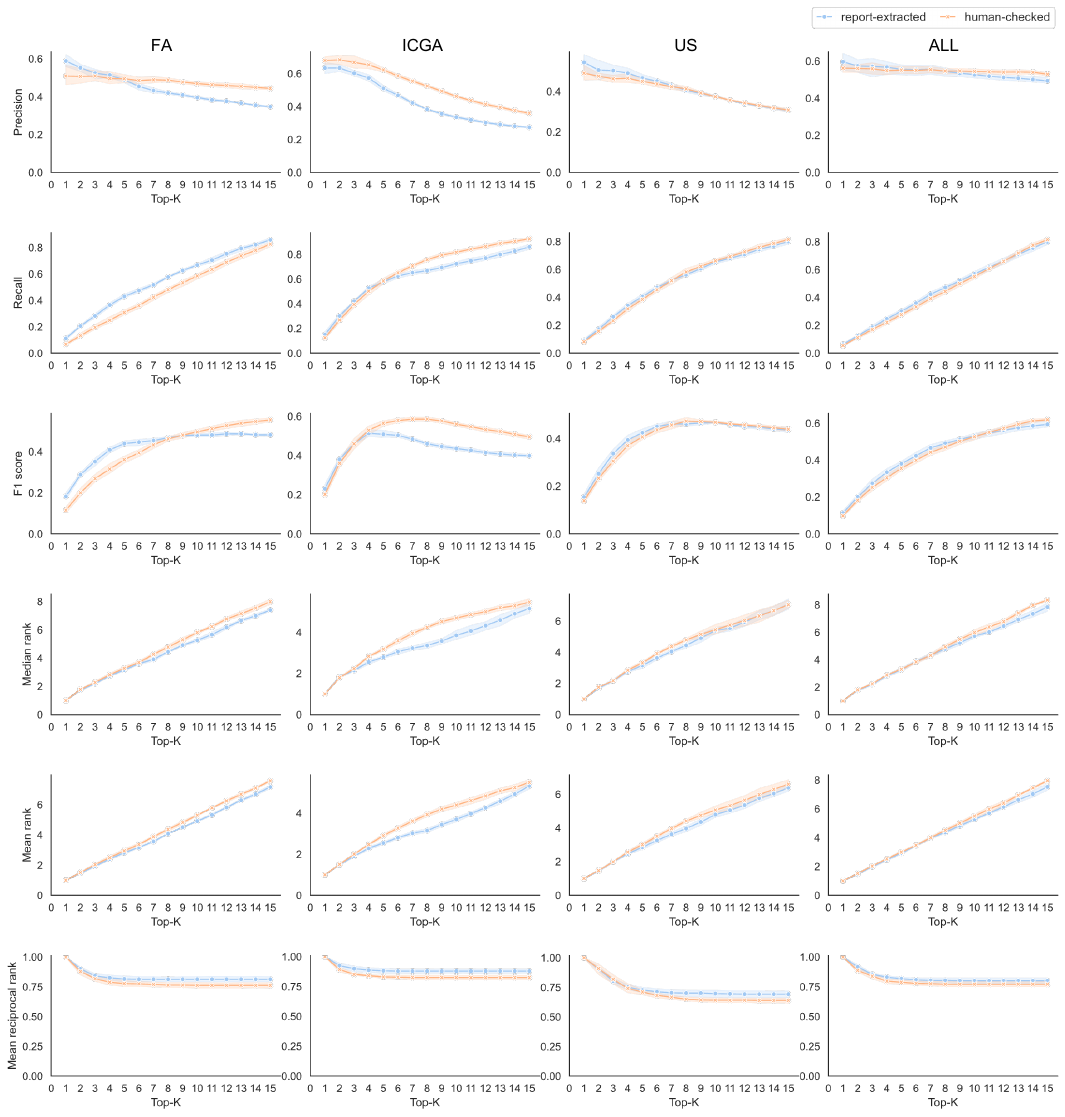}
\caption{\texttt{|} Extended details of comparative human evaluation for two concept bank generation methods across multiple modalities. Each row displays key retrieval metrics: precision@k, recall@k, and F1@k, as well as median rank@k, mean rank@k, and mean reciprocal rank@k. Each column corresponds to various data modalities: FA, ICGA, US, and ALL. The 'ALL' category represents the aggregation of top-k concepts derived from FA, ICGA and US. 
}
\label{figa1}
\end{figure}

\subsubsection{Study for the encoders across different size}
\begin{figure}[htbp]
\centering
    
    \begin{subfigure}[t]{\textwidth}
        \centering
        \includegraphics[width=\textwidth]{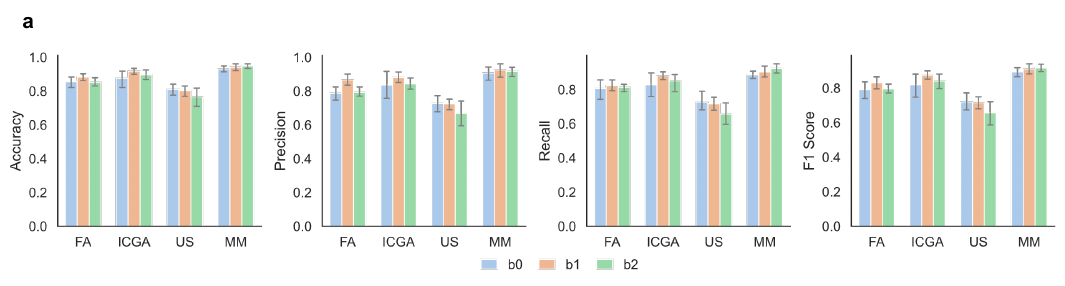}
        \phantomcaption
        \label{figa2a}
    \end{subfigure}
    \vspace{-3mm}
    
    \begin{subfigure}[t]{\textwidth}
        \centering
        \includegraphics[width=\textwidth]{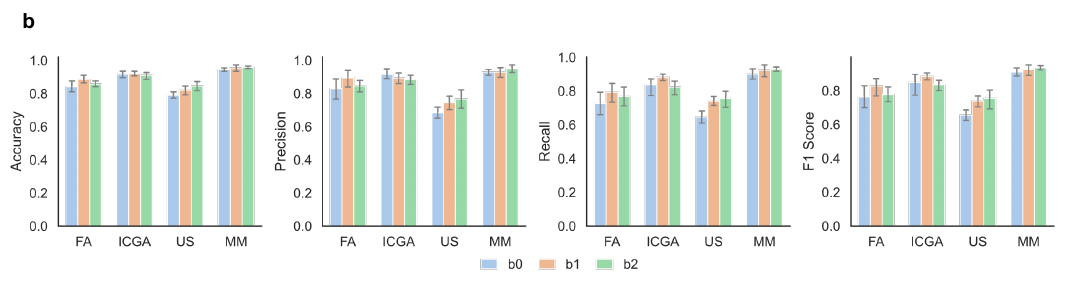}
        \phantomcaption
        \label{figa2b}
    \end{subfigure}
    \vspace{-3mm}

    \begin{subfigure}[t]{\textwidth}
        \centering
        \includegraphics[width=\textwidth]{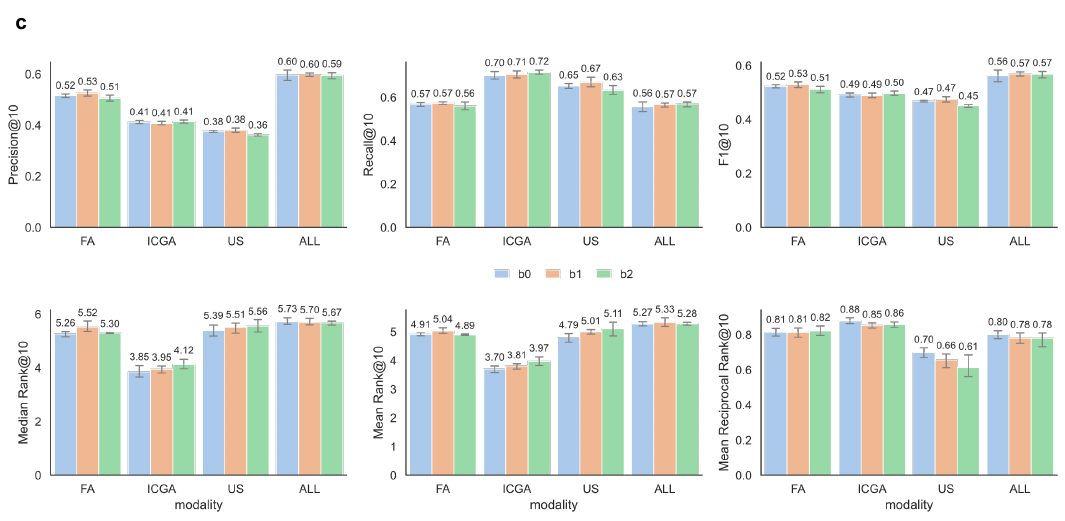}
        \phantomcaption
        \label{figa2c}
    \end{subfigure}
    
    \caption{\texttt{|} Comparative performance utilizing encoders of various sizes. (a) Comparative bar plot illustrating classification performance metrics of the pre-trained model employing different encoders (b0, b1, b2): accuracy, precision, recall, and F1 scores. Notably, the best-performing model on the validation dataset was chosen for evaluation. Metrics are represented as mean values, with error bars indicating the 95\% confidence interval based on 5-fold cross-validation on the test dataset. (b) A similar comparative bar chart is shown for the performance metrics of MMCBMs, illustrating how varying encoder sizes impact their classification effectiveness. (c) Metrics of predicted Top-k concepts of MMCBMs with different encoders (b0, b1, b2) on test dataset with k = 10. This evaluation includes precision@k, recall@k, and F1@k, as well as mean rank@k, median rank@k, and mean reciprocal rank@k.
}
    \label{figa2}
\end{figure}

To further explore the impact of encoders' size on the efficacy of pretrained models and MMCBMs. We undertook a study incorporating three variants of EfficientNet encoders—efficientnet-b0, efficientnet-b1, and efficientnet-b2—each integrated separately within the respective modalities. Fig.\ref{figa2} offers a comprehensive portrayal of the relationship between encoder size and pivotal performance metrics, depicted across three detailed comparative bar plots which focus on two key aspects: classification performance and model interpretability.

For classification performance, Fig.\ref{figa2a} and Fig.\ref{figa2b} elucidate the accuracy, precision, recall, and F1 scores. The results indicate high performance across the board with negligible differences between the encoder sizes, as evidenced by the closely clustered bars and their associated error margins, which signify a 95\% confidence interval based on 5-fold cross-validation.

Moving on to interpretability within MMCBMs, Fig.\ref{figa2c} delineates a consistent performance trend across the different encoders, without any marked disparity in the results. This section of the figure examines the precision@k, recall@k, and F1@k metrics, alongside mean rank@k, median rank@k, and mean reciprocal rank@k for predicted Top-k concepts, thereby providing an overall view of the models' interpretative capabilities.

Drawing from these insights, we have selected efficientnet-b0 as the encoder for our MMCBMs. This decision is underpinned by the encoder's ability to deliver high classification accuracy and robust interpretability, while also ensuring a more compact model with fewer parameters. The selection of efficientnet-b0 thus strikes a considered balance between maintaining performance standards and optimizing computational efficiency, which is particularly advantageous in scenarios where resource constraints are a key consideration.

\subsubsection{Ablation Study for the Concept Bank}
To further explore the efficacy of the Multi-modal Medical Concept Bottleneck Model (MMCBM), an ablation study was conducted, focusing on two critical aspects of the model: the concept extraction from diagnostic reports and the verification of these concepts. Our primary aim was to compare the impact of varying the number of reports and concepts on the performance of two distinct concept banks: \textit{the Report-Extracted Concept Bank} and \textit{the Expert-verified Concept Bank}.  Such a comparison is intended to shed light on the relative influence in the overall functioning of the MMCBM.
\medbreak
\noindent{\textbf{Report-Extracted Concept Bank}}: In our study, we first prompt the LLM to extract concepts from a set of 97 diagnose reports. Then those image-concept pairs are used directly to generate a multi-modal concept bank with CAVs. 
\medbreak
\noindent{\textbf{Expert-verified Concept Bank}}: Subsequently, the concepts derived from the LLM were meticulously reviewed by two experienced ophthalmologists. Their role was to verify the accuracy of the concept extraction, remap the relationships between images and concepts, and rectify any inaccuracies found in the initial concept set. This refined collection of concepts, validated and enhanced by medical expertise, was termed the Expert-verified Concept Bank. To streamline this verification and correction process, we developed a specialized web interface (refer to Fig.~\ref{fig6}). This interface empowered the ophthalmologists to efficiently identify and correct erroneous concepts, delete irrelevant or incorrect entries, and introduce additional concepts as necessary. 
\medbreak
\noindent{\textbf{The impact of varying the number of concepts}}.  The first part of the ablation study assessed the performance impact of varying the number of concepts extracted from these reports. For both the Report-Extracted Concept Bank and Expert-verified Concept Bank, a noticeable trend in the FA and ICGA modalities showed performance improvements as the number of concepts increased, plateauing at higher concept counts. Specifically,  for concept counts below 60, the US modality demonstrated marked improvements. Conversely, exceeding 60 concepts led to a notable decline in performance, suggesting an influx of potentially irrelevant or 'bad concepts', as indicated on our graphs. Interestingly, for the Report-Extracted Concept Bank, this trend was also apparent in the MM modality when the concept counts around 80. However, for Expert-verified Concept Bank, there is no exhibit significant decrease in performance, underscoring the importance of concept relevance and quality in the bank.

\begin{figure}[htbp]
\centering
\includegraphics[width=\textwidth]{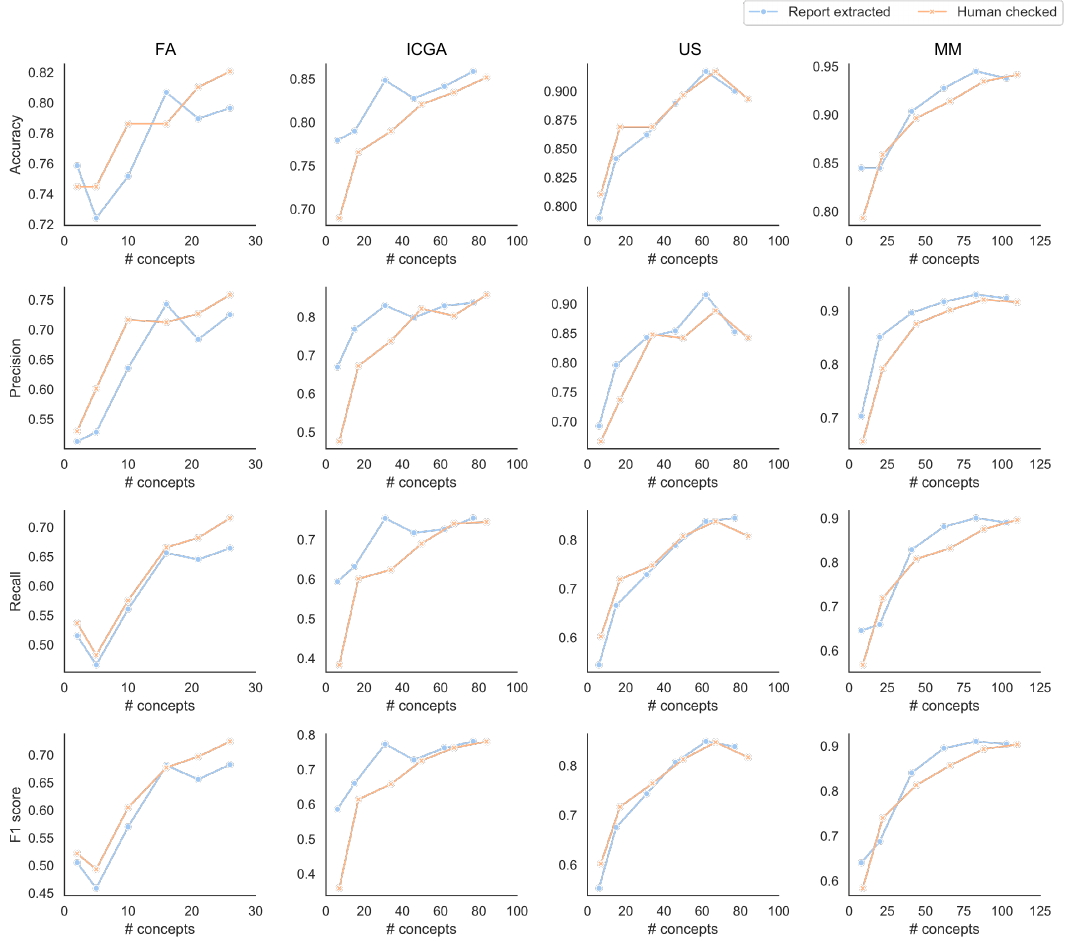}
\caption{\texttt{|} Comprehensive Ablation Study Highlighting Dependency on Concept Bank Size. The concept bank, devised leveraging reports from 100 patients with tri-modality images, forms the basis of this examination. (a) Report-extracted concept bank. The initial set of figures (first row) showcases the impact of Report Volume: This set of graphs portrays how adjusting the total number of reports impacts the model's performance metrics—precision, accuracy, recall, and F1 score. The subsequent set of figures (second row) explores the effect of Concept Count: By altering the number of disease-associated concepts, we analyze how the model's performance metrics fluctuate. This study assists in understanding the optimal number of concepts required for reliable diagnosis. (b) Expert-verified concept bank. Maintains identical settings to (a) for direct comparability.
}
\label{figa3}
\end{figure}
\begin{figure}[htbp]
\centering
\includegraphics[width=\textwidth]{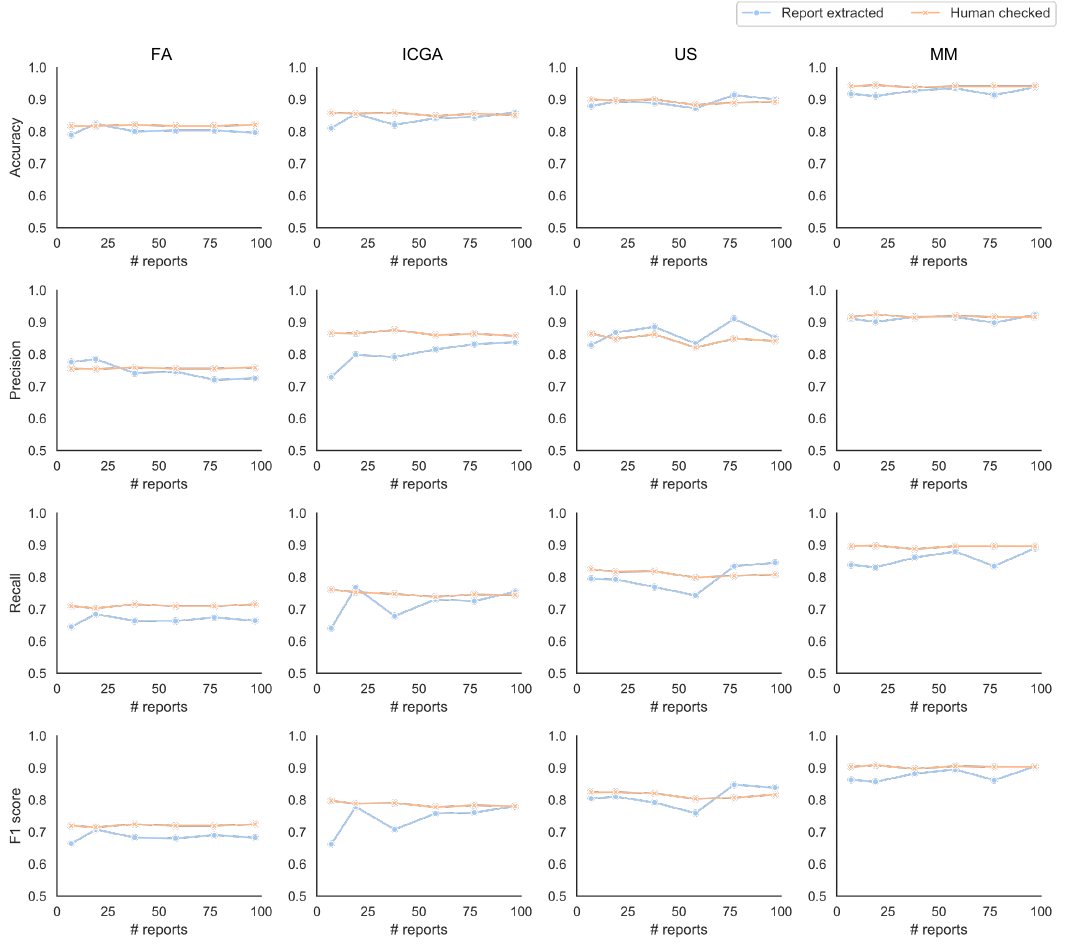}
\caption{\texttt{|} Comprehensive Ablation Study Highlighting Dependency on Number of Reports. The concept bank, devised leveraging reports from 100 patients with tri-modality images, forms the basis of this examination. (a) Report-extracted concept bank. The initial set of figures (first row) showcases the impact of Report Volume: This set of graphs portrays how adjusting the total number of reports impacts the model's performance metrics—precision, accuracy, recall, and F1 score. The subsequent set of figures (second row) explores the effect of Concept Count: By altering the number of disease-associated concepts, we analyze how the model's performance metrics fluctuate. This study assists in understanding the optimal number of concepts required for reliable diagnosis. (b) Expert-verified concept bank. Maintains identical settings to (a) for direct comparability.
}
\label{figa4}
\end{figure}

\medbreak
\noindent{\textbf{The impact of varying the number of reports}}. In the second part of our ablation study, shown in Fig.\ref{figa4}, we varied the number of reports from 7 to 97.  Contrary to our expectations, the performance metrics remained relatively stable across this range, thereby supporting the conclusion that the model's effectiveness is not substantially influenced by the quantity of reports. This observed stability in performance metrics, even with a limited number of reports, may be attributed to the comprehensive nature of the concepts contained within these reports. It appears that even a smaller set of templated reports encompasses a sufficient range of concepts. This suggests that the key determinants for model performance lie not in the sheer quantity of reports, but rather in the richness and relevance of the concepts they encompass. Notably, despite the general stability in performance across both concept banks, the Expert-verified Concept Bank consistently exhibited more robust and superior performance. This finding further emphasize the critical role of concept validity in enhancing model accuracy.

\subsection{Supplementary Methods}\label{secA3}
\medbreak
\noindent \textbf{Interpretable Predictor.} 
We trained a linear layer, denoted as the interpretable predictor $g$, to learn the preferences (weights) of specific categories for certain concepts. Then, we linearly combined these weights and used a multi-modality concept score $\boldsymbol{C}_{score}$ as the only input to obtain the final prediction. Here, $\boldsymbol{C}{\text{score}}$ represents the similarities between image features and each row within the concept bank $\mathcal{Z}_{\mathcal{C}}$. To measure the attention of the predictor on an image's concept score, we applied a sigmoid activation function to the weight and performed an element-wise multiplication between this activated weight and concept score to produce the attention matrix $W_{atten}$. Formally, considering $W$ as a learnable matrix and $\sigma$ as the sigmoid activation function, the prediction $\hat{\mathcal{Y}}$ was determined by $\hat{\mathcal{Y}}=argmax\left(\sum W_{atten}\right)$, where $W_{atten}=\boldsymbol{C}_{score} \odot \sigma(W)$ and $\odot$ represents the element-wise multiplication.

\medbreak
\noindent \textbf{Model Inference}
The Concept Bottleneck Model involves the initial transformation or mapping of the input data into a representation comprising a set of concepts, which are subsequently utilized for prediction. During the reasoning stage, the CBM model's interpretability is primarily demonstrated by its ability to provide clear explanations of how each concept contributes to the predicted results.

\medbreak
\noindent \textbf{Report Generation.}
Medical report generation (MRG) is a task that involves automatically generating a descriptive narrative report in the medical domain based on a given medical image. By utilizing our proposed interpretable predictor and predicted attention scores, we can extract multiple concepts associated with an image. This can be done by selecting either the top-k concepts or those that exceed a predetermined threshold. It is important to note that clinical diagnosis reports follow a structured format that includes components such as patient information, medical details, diagnosis, treatment recommendations, and other relevant information. Consequently, it becomes feasible to generate standardized clinical diagnosis reports by effectively combining the predicted concepts and prompting a Large Language Model (LLM) to produce comprehensive reports.

\medbreak
\noindent \textbf{Test-time intervention.}
A salient difference between CBMs and standard models is that a practitioner utilizing a CBM model can interact with it by intervening on concept predictions. This ability to intervene on concept bottleneck models enables human users to have richer interactions with them. This kind of test-time intervention can be particularly useful in high-stakes settings like medicine. For example, “correcting” the model by replacing the $j$-th concept value $\hat{c_j}$ with the ground truth value $c_j$, and then updating the prediction $\hat{\mathcal{Y}}$ after this replacement. We can qualitatively see the contribution of each concept by removing the concept and seeing the changes in the corresponding prediction's output $\hat{\mathcal{Y}}=g(\hat{\boldsymbol{C}}_{score})$, where $\hat{\boldsymbol{C}}_{score}= (\hat{c}_{\{1,...,N\}\setminus j}, c_{j})$.

\subsection{Implementation Details}\label{secA4}
In the initial assessment of our dataset, we encountered several quality issues. These included discrepancies between diagnostic reports and classification labels, images that were entirely black, blurred images with noise, and instances of data being saved twice. To address these, we implemented a rigorous data-cleaning process. Firstly, we removed all modalities containing entirely black images. Secondly, we eliminated images of poor quality, which included those with excessive blurring and noise. Thirdly, we deleted any duplicate images to ensure each data point was unique.

In our approach, we employ an early stopping mechanism to efficiently mitigate overfitting and save training time. Additionally, we perform standardization and normalization on the input images. To train the network, we set the initial learning rate to 1e-4, and the weight decay factor to 1e-2 for the Adam optimizer. The network is trained for 200 epochs, with a batch size of 8. All experiments were implemented using the NVIDIA GeForce TiTAN XP GPU and a RAM capacity of 12 GB. 

The MMCBMs underwent training and evaluation under similar conditions to previous models, with the notable difference of a learning rate set at 1e-3.

\begin{figure}[htbp]
\centering
\includegraphics[width=0.9\textwidth]{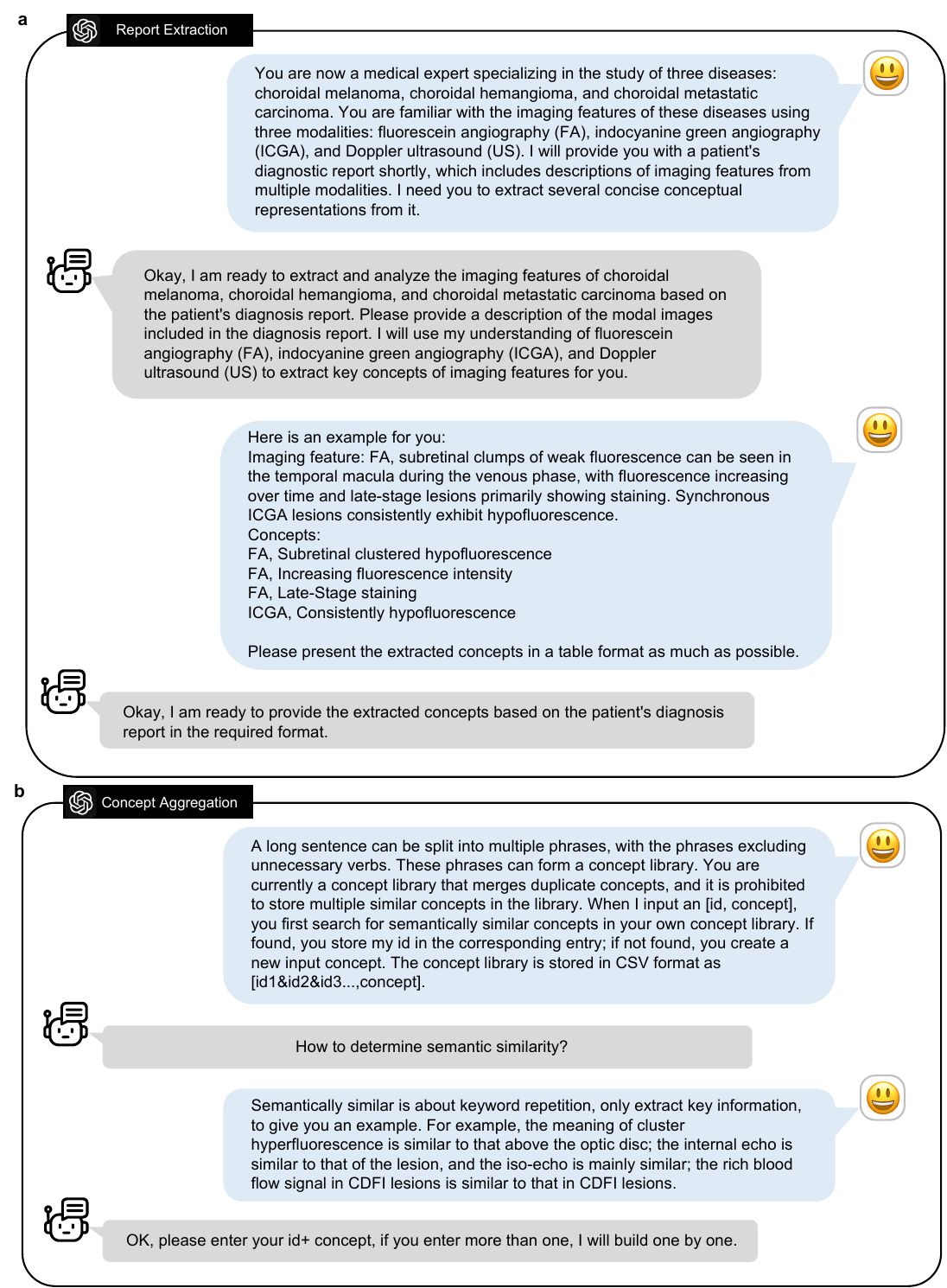}
\caption{\texttt{|} Detailed GPT-4 Prompts. (a) Concept Extraction Prompts. This part details the prompts used for extracting concepts from diagnostic reports, forming a unique concept library. (b) Concepts Aggregation Prompts. This part describes the prompts for aggregating semantically similar concepts, ensuring each concept’s uniqueness.
}
\label{figa6}
\end{figure}

\begin{figure}[htbp]
\centering
\includegraphics[width=0.9\textwidth]{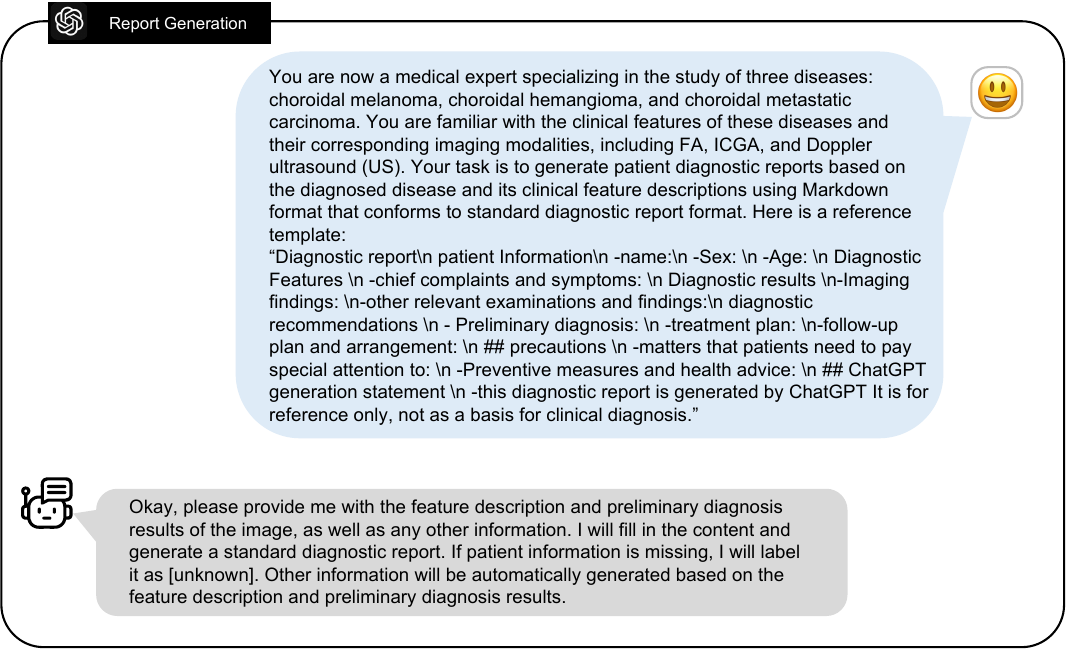}
\caption{\texttt{|} Detailed GPT-4 Report Generation Prompts. This part describes the prompts for generating patient diagnostic reports based on extracted concepts and diagnosed diseases.
}
\label{figa6-rg}
\end{figure}

\begin{figure}[htbp]
\centering
\includegraphics[width=0.8\textwidth]{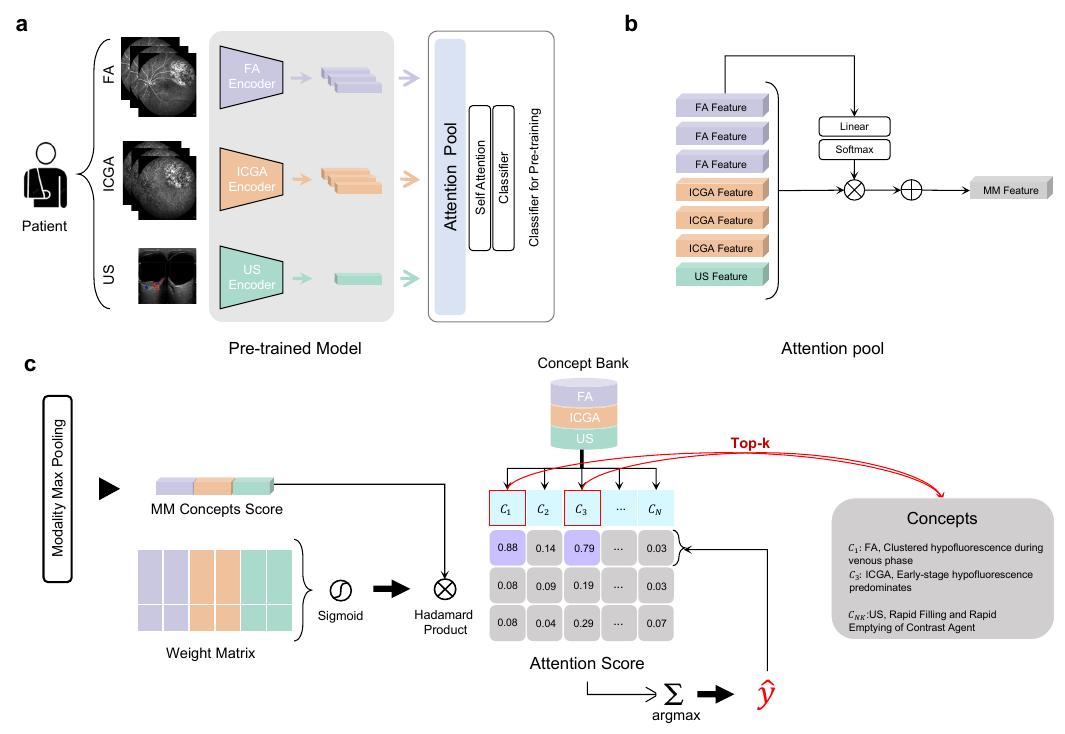}
\caption{\texttt{|} 
Details of the MMCBM Framework Module Design. (a) Illustration of the multi-modal pre-trained prediction model, where images from distinct modalities are transformed into features through dedicated encoders and merged by an attention-pooling block to determine the final classification prediction. (b) Feature Fusion Process via the Attention Pool Module. (c) Linear Layer Mechanism: a trainable weight matrix is optimized to predict tumor classes based on concept scores, with the attention score—derived from the Hadamard Product—elucidating the relationship between the current images, representing individual patients, and the respective concepts to support prediction.
}
\label{figa7}
\end{figure}

\begin{figure}[htbp]
\centering
\includegraphics[width=0.9\textwidth]{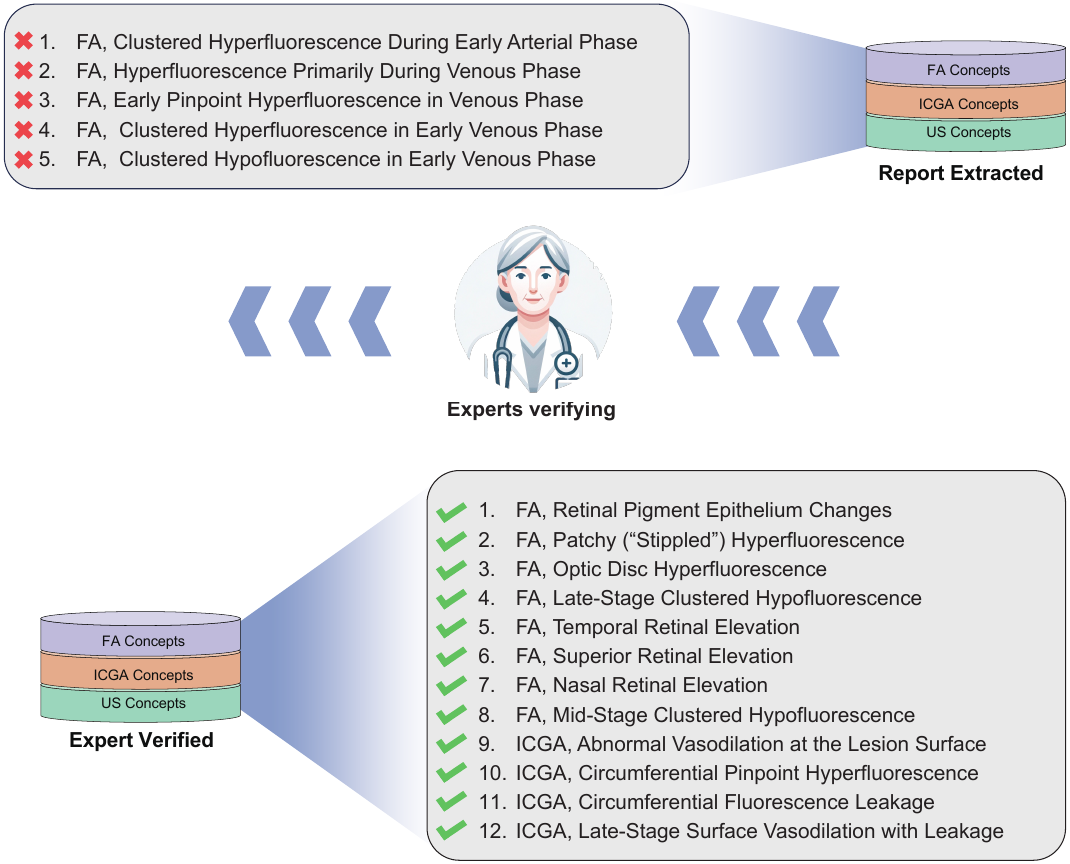}
\caption{\texttt{|} Details of the changes made by experts to the report-extracted concepts.
}
\label{figa8}
\end{figure}

\begin{figure}[htbp]
\centering
\renewcommand{\figurename}{Table}
\setcounter{figure}{0}
\renewcommand{\thefigure}{\Alph{section}\arabic{figure}}
\captionsetup{justification=raggedright, singlelinecheck=false}
\caption{Comparative Performance of Pre-trained Classifier and MMCBM on Test Dataset.}
\label{tab:taba0}
\includegraphics[width=\linewidth]{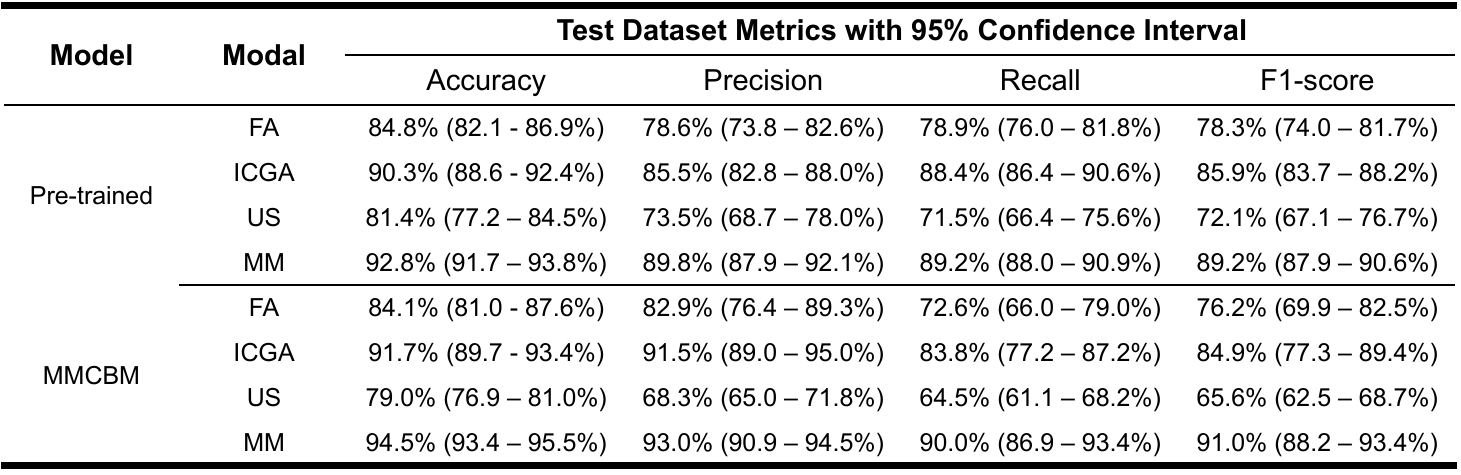}
\end{figure}

\begin{figure}[htbp]
\centering
\renewcommand{\figurename}{Table}
\captionsetup{justification=raggedright, singlelinecheck=false}
\caption{List of report-extracted concepts and the corresponding accuracy of SVMs.}
\label{tab:taba1}
\includegraphics[width=\linewidth]{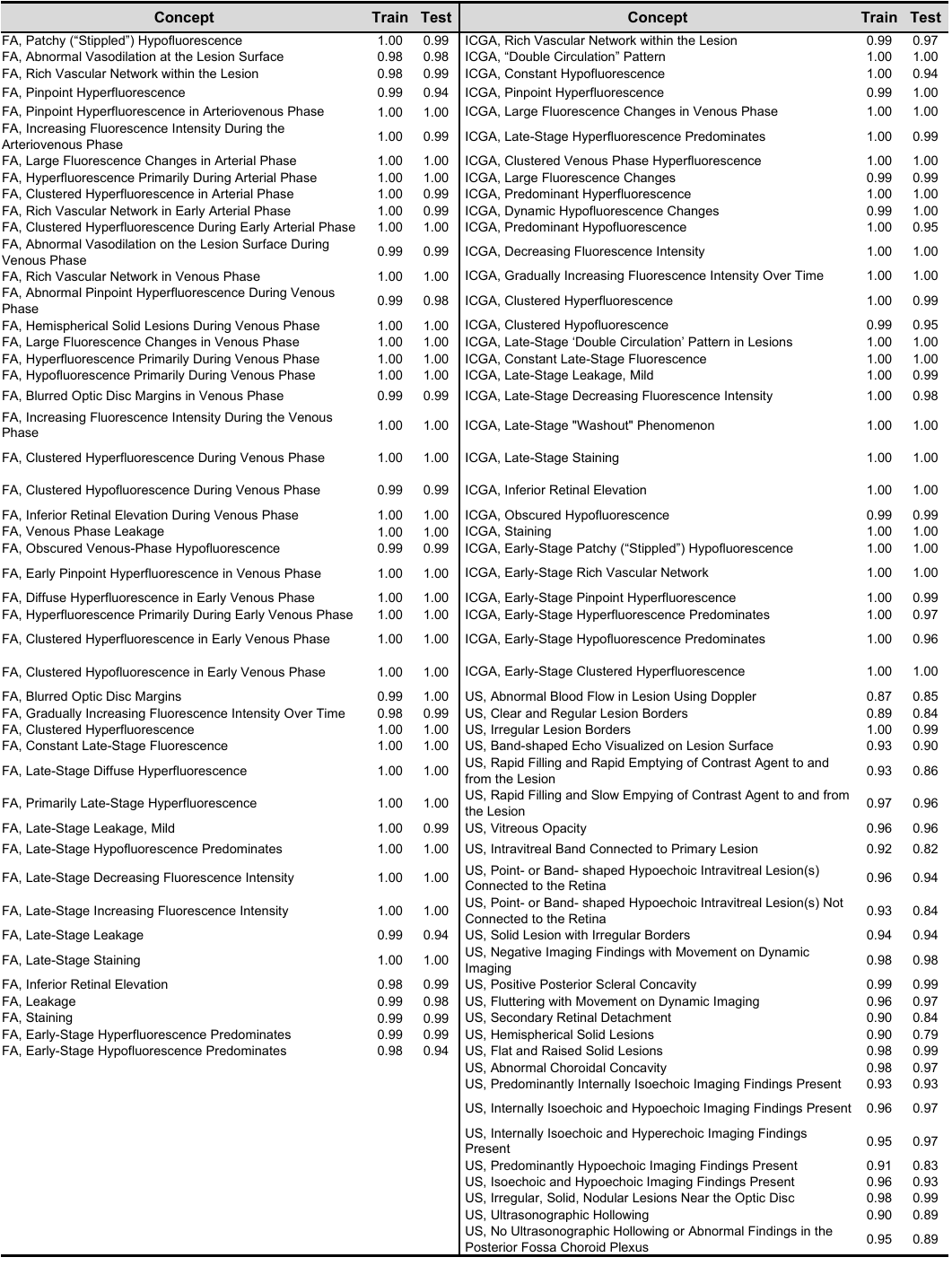}
\end{figure}

\end{appendices}
\end{document}